\documentclass{article}

\usepackage[preprint]{neurips_2025}

\usepackage[english]{babel}
\usepackage[english=american]{csquotes}
\usepackage{times}
\usepackage[T1]{fontenc}
\usepackage{wrapfig}

\usepackage{microtype}
\usepackage{amsmath}
\usepackage{amsthm}
\usepackage{amsfonts}
\usepackage{dsfont}
\usepackage{nicefrac}
\usepackage{booktabs}
\usepackage[table]{xcolor}
\usepackage{graphicx}
\usepackage{subcaption}
\usepackage{adjustbox}
\usepackage{multirow}
\usepackage{arydshln}
\usepackage{rotating}
\usepackage{array}
\usepackage{algorithm}
\usepackage{algorithmic}
\usepackage{hyperref}
\usepackage{url}
\DeclareMathOperator*{\argmax}{arg\,max}

\title{Gradient-based Learning of Simple yet Accurate Rule-based Classifiers}
\author{
	Javier Fumanal-Idocin \\
	University of Essex \\
	\texttt{j.fumanal-idocin@essex.ac.uk} \\
	\And
	Raquel Fernandez-Peralta \\
	Slovak Academy of Sciences \\
	\texttt{raquel.fernandez@mat.savba.sk} \\
	\And
	Javier Andreu-Perez \\
	University of Essex \\
	\texttt{j.andreu-perez@essex.ac.uk} \\
}

\begin{document}
	
	\maketitle
	
	\begin{abstract}
		Rule-based models are essential for high-stakes decision-making due to their transparency and interpretability, but their discrete nature creates challenges for optimization and scalability. In this work, we present the Fuzzy Rule-based Reasoner (FRR), a novel gradient-based rule learning system that supports strict user constraints over rule-based complexity while achieving competitive performance. To maximize interpretability, the FRR uses semantically meaningful fuzzy logic partitions, unattainable with existing neuro-fuzzy approaches, and sufficient (single-rule) decision-making, which avoids the combinatorial complexity of additive rule ensembles. Through extensive evaluation across 40 datasets, FRR demonstrates: (1) superior performance to traditional rule-based methods (e.g., $5\%$ average accuracy over RIPPER); (2) comparable accuracy to tree-based models (e.g., CART) using rule bases $90\%$ more compact; and (3) achieves $96\%$ of the accuracy of state-of-the-art additive rule-based models while using only sufficient rules and requiring only $3\%$ of their rule base size.
	\end{abstract}
	
	% Rest of your content goes here

\section{Introduction}

Deep neural networks (DNNs) have been used to solve complex problems in machine learning where unstructured data are available in large volumes, such as image and video \citep{lecun2015deep}. However, the use of these models is not always possible in cases where human liability is still relevant in the decision-making process, like medicine and finance \citep{arrieta2020explainable}.
Rule-based algorithms are considered one of the most trustworthy for users, as they explicitly show the patterns found and their relevance in each decision. They are also considered more faithful than other post-hoc explainable artificial intelligence (XAI) methods, which are not always reliable \citep{molnar2020interpretable,rudin2022interpretable,tomsett2020sanity}. 

By studying the rules themselves, practitioners can find additional clues about a problem, and they can also disregard the patterns found by the classifier that are contrary to existing knowledge \citep{li2024interpreting}. It is also a popular use case for rule-based inference to use the rule base as a proxy for a more complicated model, such as a deep learning model, to perform post hoc explanations \citep{Zhang2018InterpretingCV,li2024interpreting, troncoso2025new}.
One of the main research topics in rule-based classification is the trade-off between interpretability and accuracy, as the larger the number of rules, the less interpretable the model becomes. For example, ensembles of tree-based models usually perform very well, at the cost of losing the interpretation capabilities that a single tree has \citep{breiman2001random,friedman2002stochastic}. Bayesian reasoning has also been used, but it usually requires Markov chain models to train them, which is very time-consuming \citep{wang2017bayesian}. Fuzzy logic has also been considered as a tool to balance performance and rule interpretability. Fuzzy rules have been trained mainly using genetic optimization, which balances accuracy and complexity well but causes significant scaling problems \citep{alcala2011fuzzy}. It is also possible to use fuzzy rules with gradient-based optimization, but this creates fuzzy sets with no intrinsic meaning, and the training process suffers from exploding gradient issues \citep{zhang2024takagi}. Pure gradient-based approaches have also been used to train rules in classification, regression \citep{wang2023learning,zhang2023learning,yang2024hyperlogic,qiao2021learning,tuo2025interpretable} and subgroup discovery \citep{xu2024learning}. However, they can result in an arbitrarily high number of additive rules, which can only be controlled by a dense search of hyperparameters. In addition, how these models partition the state can also be hard for a user to understand, especially when the number of rules is high, as the cut points can be arbitrary and very numerous. This considerably hinders the interpretability of the model concerning a human user, which is instrumental in its deployment in real-world systems \citep{chen1994complexity}.

%One solution for this is the use of fuzzy sets for numerical variables, but they suffer from vanishing gradients and explainability issues when they are used alongside gradient descent \citep{mendel2023explainable}.

In this work, we develop a rule-based system named Fuzzy Rule-Based Reasoner (FRR). The FRR can be trained exclusively using gradient descent and combines user-defined complexity constraints (i.e., the maximum number of rules and conditions per rule) with fuzzy partitions that maintain human-readable semantics (e.g. ``low/medium/high''). In this way, we can support user-required complexity requirements while maintaining strong performance. To achieve this, our contributions are the following:
\begin{itemize}
    \item \textbf{Strict complexity control}: The proposed gradient-based rule-base learning approach allows users to establish predefined limits on the number of rules, conditions per rule, and interpretable fuzzy partitions, ensuring simplicity and clarity in the resulting explainable rule bases according to user needs.
    \item \textbf{Training Stability}: We address the non-differentiability of rule-based systems through a novel restricted addition operator and mitigate vanishing gradients with residual connections tailored for logical inference.
    \item \textbf{Flexible Rule Semantics}: FRR architecture supports both sufficient rules, where a single rule triggers a decision, and additive rules, where multiple rules contribute evidence. However, this paper focuses on using sufficient rules, as empirical studies in cognitive load theory demonstrate that they impose lower cognitive demands during interpretation than additive or hierarchical rule systems \citep{sweller2011measuring}. 
\end{itemize}

\section{Related Work}
\subsection{Rule-based models for classification}
Classical rule-based models are typically trained using heuristic or greedy algorithms due to their discrete and non-differentiable nature \citep{wei2019generalized}. While these methods are inherently interpretable, they often suffer from suboptimal convergence \citep{rudolph1994convergence} and scalability limitations, particularly in high-dimensional spaces \citep{yang2021learning}. Modern adaptations attempt to mitigate these issues by extracting rules post hoc from ensemble models like gradient boosting or random forests \citep{mctavish2022fast} or by mining frequent patterns \citep{yuan2017improved}. However, these approaches still struggle with large-scale data and often sacrifice interpretability in the process. Fuzzy rule-based systems training frequently relies on discrete optimization techniques, such as genetic fine-tuning \citep{alcala2011fuzzy}, which scale poorly with increasing rule complexity. 

Recent work has combined rule-based methods with deep neural networks to leverage the strengths of both approaches. For example, concept bottleneck networks extract relevant concepts while symbolic rules perform reasoning. Some studies fix the logic structure and focus on identifying the right concepts \citep{petersen2022,barbiero2023interpretable}, while others optimize the connections among predefined concepts \citep{vemuri2024enhancing}. However, the interpretability of such systems depends heavily on the quality of concept detection, which is difficult to evaluate. 

It is also possible to distil black-box models into rule-based ones with good results \citep{li2024interpreting}, at the expense of having to train both models.

\subsection{Gradient-based training for rule-based models} 
Due to the scalability of gradient-based optimization, several notable approaches have been proposed to extract rules from neural network classifiers. These methods often rely on the popular Straight-Through Estimator (STE) \citep{bengio2013estimating}, which is a method to approximate the derivative of non-differentiable functions like binarization. For instance, in \citep{qiao2021learning}, the authors propose a neural network architecture that uses binarization for real-valued variables and encodes the rules themselves in the neurons. Similarly, \citep{zhang2023learning} learns ``soft-intervals'' for binarization, regularized with an additional loss to avoid excessive partitioning of the feature space. However, these models often generate large numbers of rules unless trained extensively (e.g., over 1,000 epochs), compromising interpretability. This is also the problem of logic gate-based networks \citep{benamira2024truth,yue2024learning}, while they are good for automated formal verification of inference, their complexity makes it difficult for humans to interpret effectively.

Another gradient-based approach, gradient grafting \citep{wang2023learning}, combines discrete and continuous models: the continuous model guides gradient flow, while the discrete model produces layer-wise outputs. A key challenge in this gradient technique is ensuring alignment between the two models, especially in large architectures. Another problem is that the architecture proposed in \citep{wang2023learning} also tends to create large numbers of additive rules, so the reasoning mechanism is very difficult for a human being to understand.

For the case of fuzzy rule-based inference, it is possible to optimize fuzzy sets using gradient descent \citep{cui2020optimize}. However, Traditional neuro-fuzzy TSK models suffer from two key limitations when trained via gradient optimization: exponential rule growth with input dimensionality \citep{mendel2023explainable}, and interpretability loss in hierarchical architectures and their fuzzy set optimization process \citep{zhang2024takagi}. Feature selection for rules in such fuzzy systems is also non-existent, as these approaches already precompute the logic connections before optimizing the system.

Our method, the Fuzzy Rule-based Reasoner (FRR), addresses the limitations of prior crisp and fuzzy gradient-based rule learning systems by unifying trainability, interpretability, and complexity control. The FRR enables end-to-end differentiable optimization of rule systems while strictly enforcing user-defined constraints on the maximum number of rules, conditions per rule, and partitions. Furthermore, the FRR also supports sufficient rules for decisive predictions, which guarantees sparse, human-interpretable models without sacrificing accuracy. 

\section{Fuzzy Rule-Based Reasoner}
\label{sec:general_scheme}

\subsection{General scheme} 

\begin{figure*}[ht]
	\centering
	\includegraphics[width=.7\linewidth]{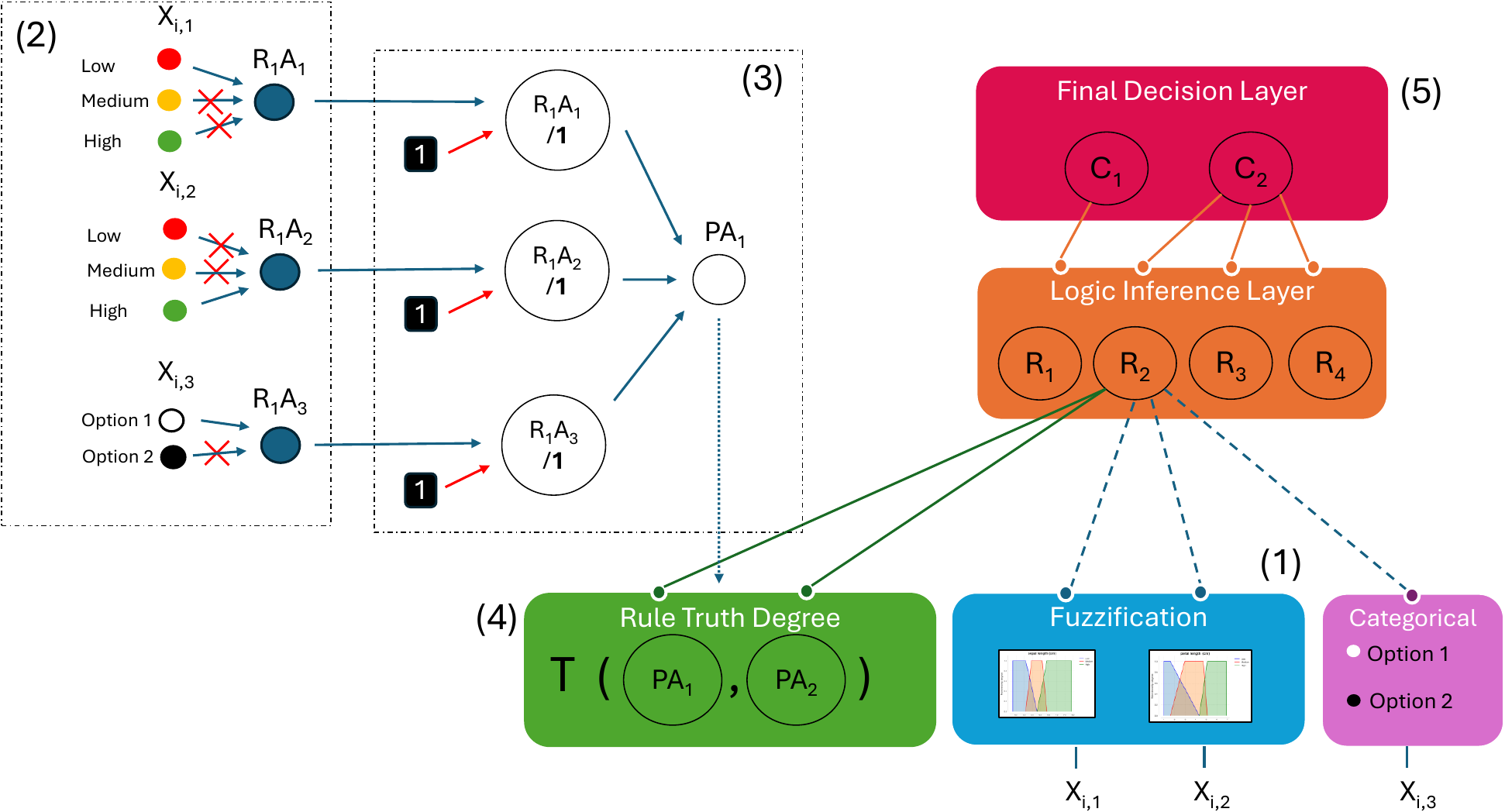}
	\caption{Fuzzy Rule-based Reasoner scheme, using an input $X_i$ with three features, four rules, and two target classes. The inference process follows: \textbf{(1)} We fuzzify the input for each real-valued variable, obtaining the degree of truth for each linguistic label. We one-hot encode categorical variables. \textbf{(2)} We forward those truth values to the logic inference layer. This layer selects the linguistic label or category for each condition. We repeat this process for the desired number of conditions per rule. \textbf{(3)} We reduce the size of the rule by determining which conditions are needed. If a condition is not needed, we substitute its truth degree with one, which is the identity element of multiplication. \textbf{(4)} We compute the truth value for each rule. \textbf{(5)} We select the output class indicated by the rule with the maximum truth degree.}
	\label{fig:scheme}
\end{figure*}

Let $\mathcal{D} = \{(X_1,Y_1),\dots, (X_N,Y_N)\}$ denote a data set with $N$ instances and $M$ features, where $X_i$ is the observed feature vector of the $i$-th instance  with $j$-th entry $X_{i,j}$ and $Y_i$ being its discrete associated target. Each feature can be either discrete or continuous, and the target is a categorical variable with $C$ classes.

The FRR is a hierarchical model with $4$ layers of matrix operations.
We will denote by $\mathcal{U}^{(l)}$ the $l$-th layer and by $u_j^{(l)}$ and $n_l$ the $j$-th node and the total number of nodes in that layer, respectively. The output of the $l$-th layer is denoted by a vector $\mathbf{u}^{(l)}$ that contains, as instances, the value of each node and $\mathbf{W}^{(l)}$ represents the connectivity matrix of layer $l$, whose structure and size are specified separately for each layer. We set a $\mathbf{W}$ matrix vector with the connectivity matrices of the different layers for each rule, but we omit this index for the sake of notational simplicity. The components of the FRR are:

\begin{itemize}
	\item \textbf{Fuzzification layer}: Transforms input features into interpretable fuzzy membership values:
	\begin{itemize}
		\item[--] \textit{Continuous variables}: Mapped to linguistic terms (e.g., ``low'', ``medium'', ``high'') via trapezoidal fuzzy sets.
		\item[--] \textit{Categorical variables}: Encoded using one-hot representation.
	\end{itemize}
	
	\item \textbf{Logic inference layer}: Computes the truth value of each rule through:
	\begin{itemize}
		\item[--] \textit{Feature selection}: Identifies relevant linguistic terms or categories for each input variable.
		\item[--] \textit{Antecedent formation}: Constructs rule antecedent from selected terms.
		\item[--] \textit{Truth degree computation}: Calculates rule activation strengths using fuzzy conjunction.
	\end{itemize}
	
	  \item \textbf{Decision layer}: Implements sufficient-rule prediction by selecting the consequent of the highest-scoring activated rule.

\end{itemize}
 Each of these layers is described below in detail and a visual scheme is shown in Figure \ref{fig:scheme}.

\subsection{Fuzzification layer for real-valued variables} 
Traditional rule-based models use dense binarization to split real-valued variables into discrete conditions. FRR instead uses fuzzy sets, enabling smoother membership degrees that reduce complexity while maintaining interpretability, as fuzzy partitions are designed to align with linguistic terms such as ``low'', ``medium'', or ``high'' \citep{zadeh1975concept}. Fuzzy logic extends classical binary logic by allowing truth values to take any real number within the range $[0,1]$ \citep{hajek2013metamathematics}. Traditional Boolean operators are then replaced with their real-valued counterparts. For example, conjunction in the FRR (the logical ``and'') is modeled using the product operator, as it is the most successful one in the literature. Other possibilities are explored as well in Appendix \ref{apx:tnorms}.

The first layer of the FRR model is a fuzzification block that transforms input observations into degrees of truth according to predefined fuzzy partitions. All numerical features are represented as fuzzy linguistic variables with at most $V$ linguistic labels. Formally, the $i$-th instance, feature $j$, and linguistic label $v$, the degree of truth is denoted by $\mu_{j,v}(X_{i,j})$. So, the output of the fuzzification layer is given by $u_{j+v}^{(1)} = \mu_{j,v}(X_{i,j})$. 

Categorical variables are represented using one-hot encoding, which can be interpreted as a degenerate form of fuzzy partitioning, where truth values are restricted to $0$ or $1$.
\subsection{Logic inference layers}\label{subsection:logic_inf_layer}

We first consider the Mamdani inference expression to compute the truth value of a fuzzy rule $r$ in an inference system
	\begin{equation}\label{eq:fuzzy_inference}
		r(X) = w_r \prod_{a \in A_r} \mu_{a}(X),
	\end{equation}
	being $w_r$ the rule weight, $A_r$ the set of conditions per rule of the rule and $\mu_a(X)$ the truth degree of the corresponding fuzzy set. %Categorical variables will not use fuzzy membership but the classical logic values $0$ or $1$.

To replicate this, we propose to separate the logic inference into two steps, which correspond to two different layers in the hierarchical model. The first step is to choose which space partitions are forwarded into the rule for each feature. With that aim, layer 2 uses a weight matrix $\mathbf{W}^{(2)}$ of size $M \times V$, i.e., the number of features per number of linguistic labels, which measures the significance of each linguistic label in the logic inference. Then, we only forward the linguistic label with the highest weight value per each feature. In accordance, the output of the second layer is
\begin{equation}\label{eq:u2}
u_j^{(2)} =  \sum_{v=1}^V f(W^{(2)}_{j,v}) W^{(2)}_{j,v}u_{j+v}^{(1)} = \sum_{v=1}^V f(W^{(2)}_{j,v}) W^{(2)}_{j,v}\mu_{j,v}(X_{i,j}),
\end{equation}
where $f$ is an indicator function that, given an instance $M_{i,j}$ of the $i$-row of a certain matrix $\mathbf{M}$, returns 1 if that instance has the highest value in the row, i.e.,
\begin{equation}\label{eq:f}
    f(M_{i,j})=\left\{
    \begin{array}{ll}
        1 & \text{if } j = \argmax_{k} M_{i,k},\\
        0 & \text{otherwise.} 
    \end{array}
    \right.
\end{equation}

The second step for performing the logic inference is to choose which features are selected as part of the antecedent of the rule. To set the number of conditions per rule to a fixed size $A$, in the third layer, we use a weight matrix $\mathbf{W}^{(3)}$ of size $A \times M$, 
 which quantifies the relevance of each feature per condition in the rule. The contribution of the $k$-th condition is given by the following equation:
\begin{equation}\label{eq:slot_ant}
    A_k=\sum_{j=1}^M f(W^{(3)}_{k,j}) W^{(3)}_{k,j}u_{j}^{(2)},
\end{equation}
where $f$ is the function defined in Eq. (\ref{eq:f}). Consecutively, the degree of truth of each rule is computed as the product of all conditions' contributions and is the output of the third layer
\begin{equation}\label{eq:u3}
    u^{(3)} = \prod_{k=1}^A A_k =
    \prod_{k=1}^A \sum_{j=1}^M f(W^{(3)}_{k,j}) W^{(3)}_{k,j}u_{j}^{(2)}.
\end{equation}

To have a valid fuzzy logic inference process, we need to keep the domain of Eq. (\ref{eq:u3}) inside $[0,1]$. Since the input values are already in that range and only the multiplication operation is used, it is sufficient to make sure that the weights are also in that range to keep everything in $[0,1]$. To do so, each weight in Eq. (\ref{eq:u3}) is transformed using the softmax function
\begin{equation}\label{eq:softmax}
\tilde{W}^{(l)}_{i,j}={\frac {e^{ W^{(l)}_{i,j}/\alpha}}{\displaystyle \sum _{m}e^{W^{(l)}_{i,m}/\alpha}}},
\end{equation}
where $\alpha$ is a positive real number called the temperature parameter and controls the sharpness of the distribution. Tuning this parameter during training can be used to approximate an indicator function \citep{balin2019concrete}. We set in our experiments $\alpha=0.1$, so that Eq. (\ref{eq:softmax}) becomes a better approximation of Eq. (\ref{eq:f}), which will be relevant when computing the gradients of the model.

\subsection{Final decision layer}

This module computes the predictions of the FRR given the truth degrees of the rules. In this layer, we consider the matrix $\mathbf{W}^{(4)}$ of size $R \times C$ where $R$ is the number of rules and $C$ is the number of classes of the target variable. Then, each weight $W_{s,c}^{(4)}$ corresponds to the score that rule $r_s$ gives to class $c$. Taking into account the truth degrees of each rule provided by the previous layer,  we can compute the final outcome, denoted in general by $u^{(4)}_c$: for a fixed class $c$ we consider the set of rules whose maximum score is assigned to $c$, and then we consider the value of the highest score per truth degree as output
\begin{equation}\label{eq:sufficient}
	FRR(X_i)_c^{suf} = \max_{s \in \{1,\dots,R\}} f(W_{s,c}^{(4)})W_{s,c}^{(4)} r_s(X_i).
\end{equation}

\subsection{Making the model parsimonious} \label{sec:silencer}

While Section \ref{subsection:logic_inf_layer} establishes a fixed upper bound for conditions per rule, we further reduce rule length through a competitive cancellation mechanism. This allows the model to prune unnecessary conditions dynamically during training without compromising performance. To do this, we make each condition ``compete'' with a null element that will make the system ignore it if it is not useful. To do so, for each condition in the rule, we consider instead a linear combination of that condition's truth value and a static weight:
\begin{equation}\label{eq:antecedent_silencer}
	u_k = \prod_{k=1}^A (\alpha_{k,1}f(\alpha_{k,1})A_k + \alpha_{k,2}f(\alpha_{k,2})),
\end{equation}
where $\alpha_{k,1}, \alpha_{k,2} \in \mathbb{R}$. So, when $\alpha_{k,2} > \alpha_{k,1}$, $A_k$ is ignored to compute the rule truth value.

\subsection{Extracting the rules from the model}
Once trained, we can recover the rules obtained by the system by following ``the paths'' in the FRR.  For each rule, we select the features that have the biggest weights according to $\mathbf{W}^{(3)}$ that were not canceled in Eq. (\ref{eq:antecedent_silencer}). Then, for each feature we select the linguistic label according to $\mathbf{W}^{(2)}$ and finally, in the decision layer, we choose the consequent according to $\mathbf{W}^{(4)}$ considering Eq. (\ref{eq:sufficient}).

\section{Training the FRR}
\label{sec:training}

The FRR presents three optimization challenges inherent to the logic inference scheme chosen (see Appendix \ref{apx:diff} for a complete differentiation study). In the following, we discuss how we solved them.

\subsection{Non-differentiable Indicator Function}

The function $f$ in Eqs.~(\ref{eq:u2}) and~(\ref{eq:u3}) uses an $\argmax$ operation, which is non-differentiable, so we need to approximate it. This approximation in the literature is typically the derivative of the identity function (STE) or another smooth function that behaves similarly to the original \citep{yin2018}. We will use the former, as there is no guarantee that other functions will work better than the identity function \citep{schoenbauer2024}.
%A complete differentiation study can be found in Appendix \ref{apx:diff}.

\subsection{Gradient Sparsity in Rule and Antecedent Selection}

The hard rule selection creates gradient sparsity, as only the maximal weight receives updates in Eqs.~(\ref{eq:u2}) and~(\ref{eq:u3}). To improve exploration in training, we propose a relaxed version of the indicator function during training time that will also keep the standard behaviour of the model at test time. Let $\mathbf{M}$ be a matrix of size $n \times m$, we define: 
\begin{equation}\label{eq:fbeta}
	f_\beta(M_{i,j})=\left\{
	\begin{array}{ll}
		\frac{1}{1+\beta(m-1)} & \text{if } j = \argmax_{k} M_{i,k},\\[0.2cm]
		\frac{\beta}{1+\beta(m-1)} & \text{otherwise,} 
	\end{array}
	\right.
\end{equation}
where $\beta \in [0,1]$ is a hyperparameter of the model. In this way, when $\beta>0$, the additive nature of the summations is restricted but not completely ignored. When $\beta=0$, Eq. (\ref{eq:fbeta}) is equivalent to a hard indicator function. Since $\sum_{j=1}^m f_{\beta}(M_{i,j})=1$ always holds regardless of the value of $\beta$, the output will not change the scale of the input values. 

To set the value $\beta$ during training, we start with a maximum value (usually $1$) and gradually decrease it to a minimum value (usually $0$). This allows the model to explore freely with larger gradient flows initially and resemble the final inference behaviour in later epochs.

\subsection{Vanishing Gradient in Rule Inference}

The product in Eq.~(\ref{eq:u3}) used to compute the truth value of the antecedent of the rule is composed of $A$ terms in $[0,1]$, which leads to values close to 0 and creates exponential gradient decay with increasing rule complexity. To solve this, we propose two complementary approaches:

\begin{itemize}
	\item \textbf{Root-normalized activation:} we modify the product computation using $P(x) = \sqrt[n]{x}$, so that the values close to $0$ will are bigger:
	\begin{equation}\label{eq:P}
		\tilde{u}^{(3)} = \prod_{k=1}^A P(A_k).
	\end{equation}
	
	\item \textbf{Residual connections:} we consider another connection between the rules output and their antecedent values, similar to residual connections \citep{he2016deep}, to tackle the problem with the vanishing gradient in Eq. (\ref{eq:u3}):
	\begin{equation}
		%\tilde{u}^{(3)} = u^{(3)} + \gamma \sum_{j=1}^M u_j^{(2)}.
		\tilde{u}^{(3)} = u^{(3)} + \gamma \sum_{k=1}^AA_k.
	\end{equation}
	\noindent The multiplying constant $\gamma$ starts at $0.1$ and decreases linearly with the number of epochs passed in the training process, reaching $0$ at the end and in the inference process.
\end{itemize}

\section{Experiments} \label{sec:exps}

\subsection{Experimental settings}

We took $40$ datasets of different sizes, all of which are very common in studying classification performance. They range from $80$ to $19020$ samples and from $2$ to $85$ for different numbers of features (complete dataset specifications are provided in Appendix \ref{sec:datasets}).

As a classification metric, we use the standard accuracy. We use a 5-fold evaluation to obtain more reliable results than a traditional $80/20$ split. To determine statistical differences between different classifiers, we use Friedman Test and Post-Hoc Nemenyi \citep{demvsar2006statistical}.

While rule base complexity can be assessed through both structural and semantic measures \citep{MENCAR20114361, gacto2011interpretability}, we focus on structural metrics for their objectivity and demonstrated correlation with human interpretability. Traditional size metrics, however, overlook critical factors like condition reuse (e.g., in decision trees) that enhance comprehensibility. So, besides rule base size, we also reported the average number of unique logical conditions per rule base.

%The fuzzy partition itself can be a limiting factor in the performance of the FRR. To solve this, the parameters of the trapezoids can be optimized through gradient descent as well,  because similarly to ReLU, trapezoidal functions are differentiable everywhere except by four points, which does not cause issues during training.  However, we are not warranted to obtain a reasonable result in that case i.e. lower values in the input space should always have a lower membership value to ``high'' and ``medium'' than to ``low''.

 \subsection{Performance study and comparison with other explainable and standard classifiers}

\begin{table*}[ht]
		\caption{5-fold results for all the datasets considered. }
	\begin{adjustbox}{width=\linewidth}
	\centering
	\begin{tabular}{l|cccccccccc|c|c}
		\toprule
		& \multicolumn{5}{c}{Sufficient Rule-based} & \multicolumn{3}{c}{Tree-based} & \multicolumn{2}{c}{Additive Rule-based} & & \\
		\cmidrule(lr){2-6} \cmidrule(lr){7-9} \cmidrule(lr){10-11}
		Method              & FRR   & FGA    & DRNet & DINA   & RIPPER & CART   & C4.5   & GOT   & SIRUS  &   RRL      & LR & GB \\
		\midrule                                                                                                             
		Accuracy            & 79.51 & 70.46  & 56.08 & 54.99  & 75.22  & 81.06  & 79.99  & 76.91 & 82.17  &  81.99     & 82.12 & 86.04 \\
		\hdashline                                                                                                                   
		Number of Rules     & 13.77 & 7.12   & 24.04 & 18.48  & 16.04  & 39.75  & 131.92 & 5.23  & 286.71 & 99.35      & --    & -- \\
		Conditions/Rule     & 1.94  & 2.23   & 6.37  & 4.59   & 1.96   & 5.75   & 8.10   & 2.27  & 2.90   & 8.85       & --    & -- \\
		Rule base Size       & 26.71 & 15.87  & 153.13&  84.82 & 31.43  & 228.56 & 1068.55& 11.87 & 831.45 & 879.24     & --    & -- \\
		Unique Conditions   & 10.78 & 10.52  & 16.26 & 9.18   & 21.30  & 34.72  & 68.56  & 11.00 & 357.05 & 125.16     & --    & -- \\
		%Global complexity   & 3.62  & 3.27   & 5.13  & 4.54   & 3.96   & 5.57   & 7.03   & 3.12  &  7.08  & 6.91       & --    & --  \\
		\bottomrule
	\end{tabular}
\end{adjustbox}
	\label{tab:cmp_results}
\end{table*}

For a comprehensive evaluation, we compared against three categories of baselines: gradient-optimized and traditional rule-based systems, tree-based models, and standard non-rule-based classifiers.

For rule-based methods we tested a total of 6 different methods besides the FRR: an equivalent Fuzzy Rule Based classifier using genetic fine tuning \citep{fumanalex2024} (FGA); extracting and pruning rules from random forests \citep{benard2021interpretable} (SIRUS); extracting rules from a Neural network that used discretized inputs (DRNet) \citep{qiao2021learning}; Repeated Incremental Pruning to Produce Error Reduction (RIPPER) \citep{cohen1995fast}; Rule-based  Representation Learning (RRL), which uses a custom neural architecture and gradient grafting for build rules \citep{wang2021scalable};  DiffNaps, which consists on using  a neural network to reconstruct relevant patterns combining a classification and reconstruction loss \citep{walter2024finding} (DINA). The last three of them are gradient-based rule learning algorithms.

The tree-based methods are: classification trees constructed using Classification and Regression Trees (CART) methodology \citep{timofeev2004classification}, C4.5 \citep{quinlan1993c4} and Generalized Optimal Sparse Decision Trees with heuristic binarization techniques \citep{mctavish2022fast} (GOT).

We also show results for two models that are not rule-based: a Logistic Regression (LR) and Gradient Boosting (GB) \citep{friedman2001greedy}, which is considered non-interpretable.

Table \ref{tab:cmp_results} shows the results for all the datasets and all the classifiers tested.

\subsection{FRR performance}

 \begin{figure}
 	\centering
\begin{subfigure}{0.32\linewidth}
	\centering
	\includegraphics[width=\linewidth]{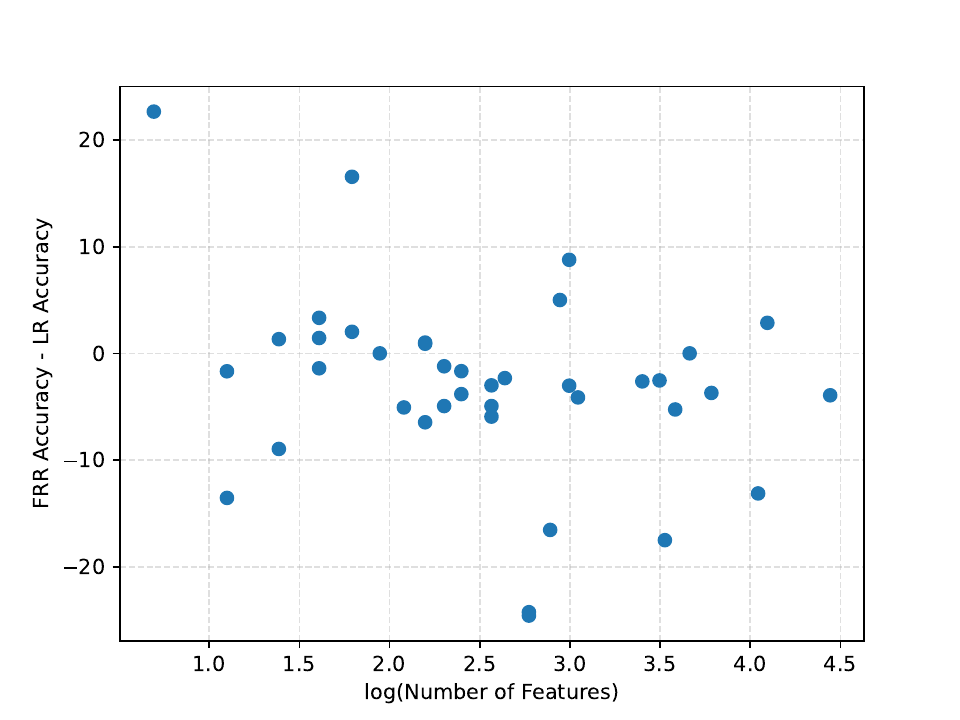}
	\caption{}
\end{subfigure}
\hfill
\begin{subfigure}{0.32\linewidth}
	\centering
	\includegraphics[width=\linewidth]{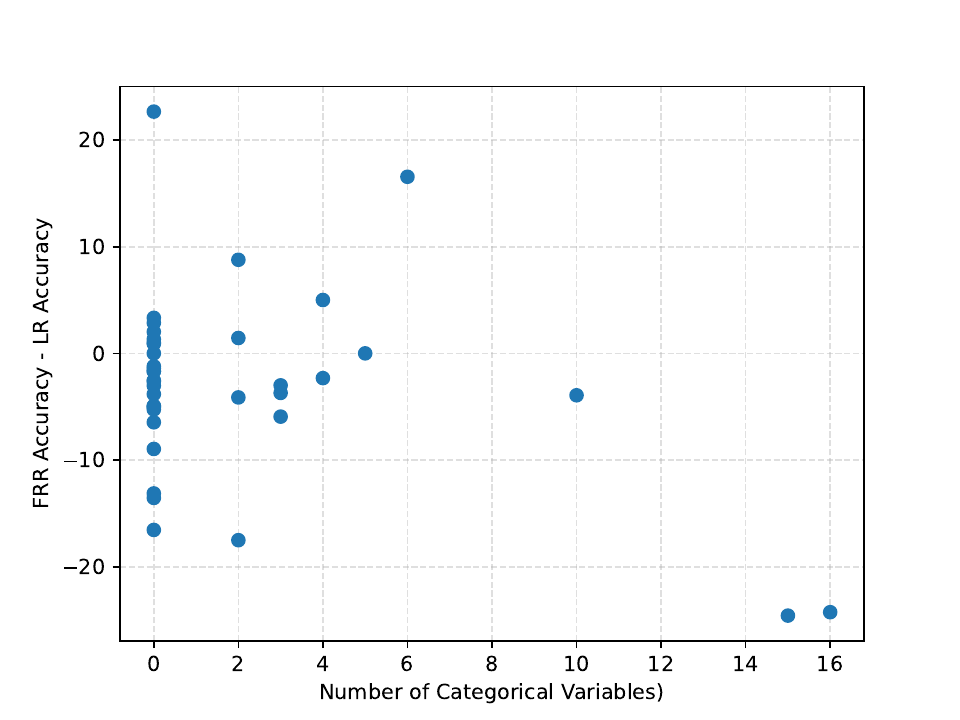}
	\caption{}
\end{subfigure}
\begin{subfigure}{0.32\linewidth}
	\centering
	\includegraphics[width=\linewidth]{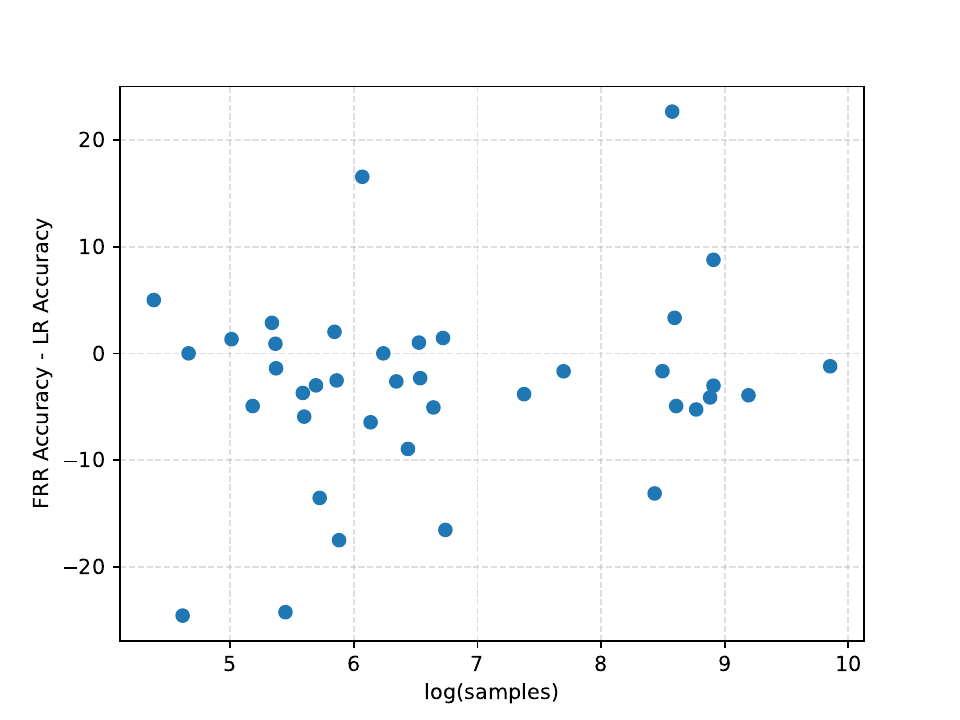}
	\caption{}
\end{subfigure}
\hfill
\caption{Performance study of the FRR according to the number of features in the dataset (a), the number of categorical features (b) and number of samples (c).}
\label{fig:main_scalability}
\end{figure}

The performance of the Fuzzy Rule-based Reasoner (FRR) was evaluated across datasets with varying characteristics, as illustrated in Figure \ref{fig:main_scalability}. In all the experiments, we fixed the total maximum number of rules to 15 and the conditions in antecedent to 3. The results highlight the model's robustness and adaptability under different conditions, including feature dimensionality, data types, and sample sizes. To measure this, we compare the result obtained in each dataset with the FRR with respect to the LR, which allows to take into consideration the difficulty for each dataset. 

We found that the FRR maintains consistent accuracy across datasets with 2 to 85 features, demonstrating its scalability. A slight performance degradation is observed for very high-dimensional datasets (e.g., coil2000, with 85 features), likely due to sparsity constraints limiting the coverage of learned rules. This suggests that while FRR handles moderate feature spaces effectively, extremely high-dimensional problems may require additional mechanisms to effectively ensure rule coverage.

Datasets with up to six categorical variables exhibit good behaviour, while those composed with only categorical (i.e. housevotes, zoo) have significantly worse performance. This could be due to the nature of such datasets, where additive contributions work better, or when very long sequences of conditions are more suited than sets of rules, which is why additive models and trees performed well in comparison. FRR’s accuracy also remained stable with sample size. However, smaller datasets (e.g., hepatitis, with 80 samples) show higher variance in performance, which shows the need for sufficient data to learn robust and generalizable rules. For a concrete example of the output rule base of the FRR see Appendix \ref{apx:example}.

\subsection{Accuracy and complexity comparison}

\begin{figure}
	\centering
	\includegraphics[width=.7\linewidth]{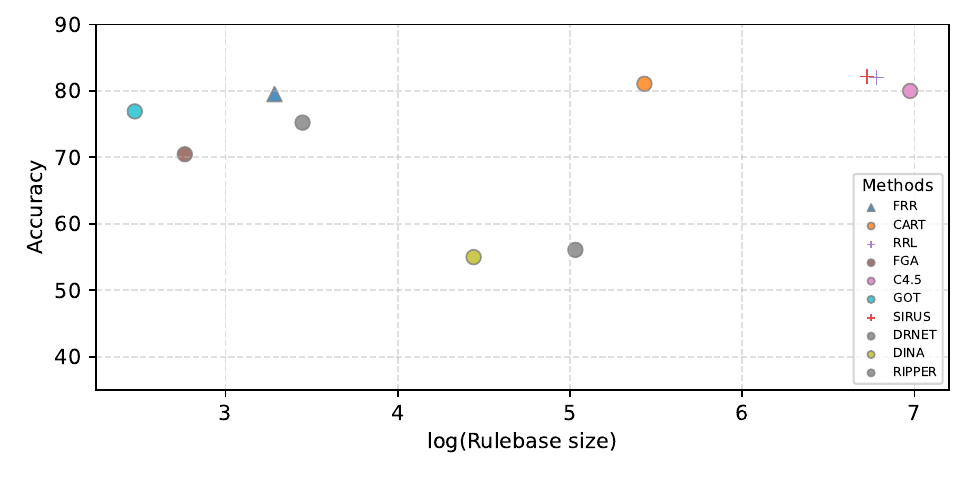}
	\caption{Relation between rule-based classifier complexity and their accuracy. The logarithmic transformation of the complexity axis serves the purpose of normalizing the scale difference between small and large rule bases, ensuring that complexity differences between 2 vs. 3 rules are treated proportionally as to 150 vs. 170 rules. Sufficient rule-based classifiers are marked with a circle, except for the FRR, and additive ones with a plus symbol.}
	\label{fig:complexity_accuracy}
\end{figure}

The relationship between complexity and accuracy for all rule-based classifiers is shown in Figure \ref{fig:complexity_accuracy}. First, we noted that observed that none of the rule-based classifiers obtained a comparable performance to the Gradient Boosting classifier, and only the additive models, SIRUS, with an average accuracy of $82.17\%$, and RRL, with $81.99\%$, obtained statistically equivalent performance to the LR. We can deduce from this that some degree of additive reasoning is relevant to some classification problems. Moreover, these rule-based classifiers are not only additive-based but also very complex in terms of size, which induces us to think that most of the rules learned by them might not be very useful, especially if we consider how big they are compared to the other rule-based classifiers considered. For example, the FRR obtained a $79.51\%$ average accuracy, which is $96\%$ of what SIRUS obtained, by using a rule base that was only $3\%$ the size of the SIRUS one.

Regarding these sufficient rule classifiers, the FRR was the best one and was deemed statistically superior to the second best, RIPPER ($75.22\%$). The gradient-based methods, DRNet and DINA, significantly underperformed compared to the rest of the models. We believe that the reason for this is the fact that these models were designed for very large problems, and some of the tips that the authors gave in the paper, i.e., using a large batch size, did not work well in this test. These methods also demonstrated high sensitivity to hyperparameter configurations, which was not feasible to optimize for all datasets in our study.

Concerning tree-based models, the best performing one was CART, with $81.06\%$, followed by C$4.5$, with $79.99\%$. Following them, GOT obtained a $76.91\%$, while being significantly less complex than the other two. Compared to the FRR, C$4.5$ was deemed statistically equivalent and CART was deemed superior. However, the FRR proved again its superiority complexity-wise: its rule base size was only $11\%$ of the CART algorithm and $2.5\%$ of the C$4.5$.

%\begin{wrapfigure}{lht}{0.5\textwidth}
%\end{wrapfigure}

\subsection{Ablation study}

\begin{wraptable}{r}{0.5\textwidth}

%\begin{figure}
\centering
\begin{subfigure}{0.48\linewidth}
    \includegraphics[width=\linewidth]{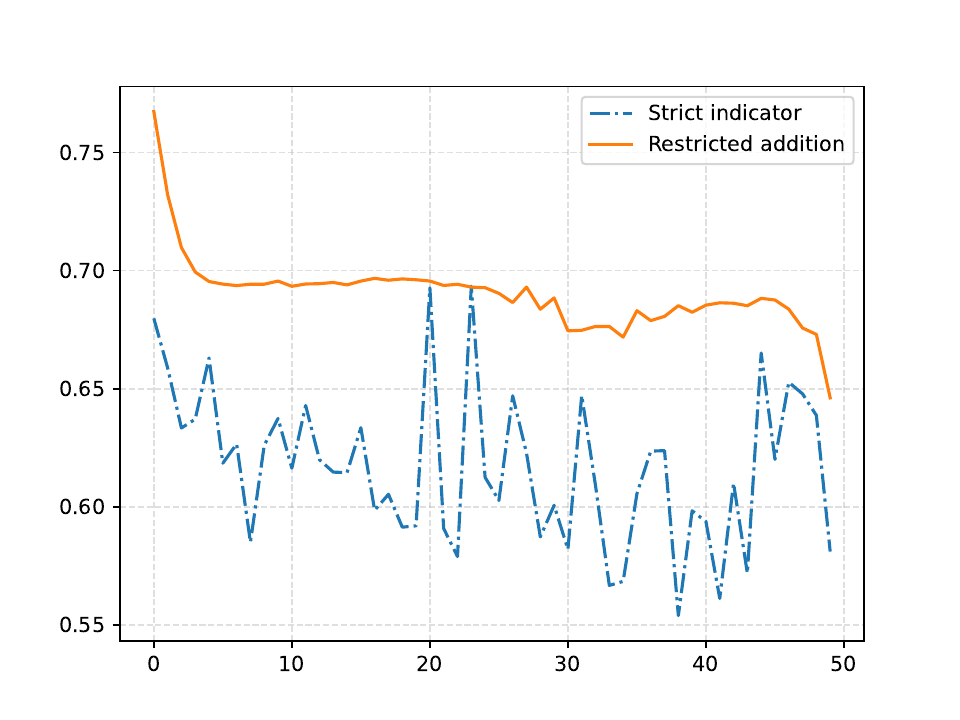}
\end{subfigure}
\begin{subfigure}{0.48\linewidth}
    \includegraphics[width=\linewidth]{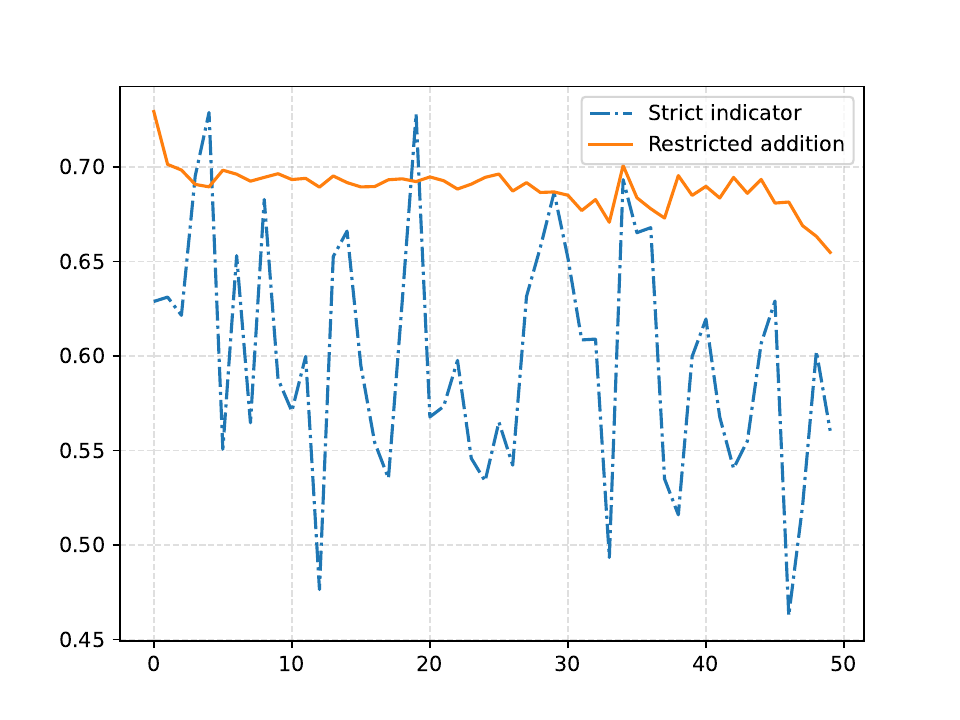}
\end{subfigure}
	\caption{Loss evolution of the Pima and Australian datasets with and without restricted additions.}
	\label{fig:ra_ablation}
%\end{figure}
\end{wraptable}
%Our ablation study demonstrates the critical contributions of FRR's key components
%In our experiments, we employed the FRR model with residual connections and restricted additions. To evaluate the contribution of these components, we conducted an ablation study by systematically removing each one. 
%Without residual connections, the accuracy reduced by an average $3.4\%$ (Friedman $p$ < 0.01) over the baseline ($76.10\%$).
Our ablation study quantitatively confirms FRR's design. Removing residual connections results in an average reduction of $3.4\%$ in accuracy (Friedman $p$ < 0.01) over the baseline ($76.10\%$). 
Restricted addition significantly stabilizes training (see Figure \ref{fig:ra_ablation}), reducing loss volatility compared to the unstable indicator baseline. This translates to consistent accuracy gains. For the two examples shown in Fig. \ref{fig:ra_ablation}: the restricted addition model on the Pima dataset achieved $+7.46\%$ ($72.79\%$ vs. $65.33\%$) and $+7.29\%$ on the Australian dataset ($83.91\%$ vs. $76.62\%$) over the baseline.

\section{Conclusions and future work}

We introduce the Fuzzy Rule-based Reasoner (FRR), a novel explainable classifier that learns interpretable rules through gradient-based optimization. The FRR comes with a big advantage over other rule-based methods: the user can set the maximum number of rules and the length of their antecedents. This avoids common problems in real-life applications where rule bases demand an excessive cognitive load for human stakeholders to understand. The FRR also presented some challenges, as rule learning inherently involves non-differentiable functions, and it also showed vanishing gradient problems, which we studied and mitigated. We probed in the experimentation that the FRR can obtain a very good balance in terms of accuracy and complexity using sufficient rules, surpassing the performance of the other sufficient gradient-based rule learning methods, and achieving similar performance to tree-based classifiers with lesser complexity. Going forward, we aim to use the FRR inside bigger deep learning frameworks with symbolic outputs, so that it can offer rule-based explanations of these models' predictions within the same gradient flow. We also intend to study the relationship between the rules learned by the decision layer and epistemic and aleatoric uncertainty.

\section*{Acknowledgment}

Raquel Fernandez-Peralta is funded by the EU NextGenerationEU through the Recovery and Resilience Plan
for Slovakia under the project No. 09I03-03-V04- 00557.
Javier Fumanal-Idocin research has been supported by the European Union and the University of Essex under a Marie Sklodowska-Curie YUFE4 postdoc action.

\bibliographystyle{kr}
\bibliography{kr-sample}

\newpage
\appendix
\section{Computing the fuzzy partitions for real-valued features} \label{sec:fuzzyfication_details}

For our fuzzy partitions, we use trapezoidal memberships as they are easier to interpret than Gaussians. We build them using a common setup in the literature where $3$ linguistic labels are defined, which should be constructed so that they can be mapped to the concepts of ``low'', ``medium'' and ``high'', which them easy to interpret to the final user \citep{gacto2011interpretability}.  We do so by setting their parameters according to the following quantile distribution:

\noindent Let $\mathbf{X}$ be a matrix of shape $(n, m)$ where $n$ is the number of samples and $m$ is the number of variables.
\begin{enumerate}
	\item Compute quantiles:
	Define quantile percentages: $q_i = \{0, 20, 40, 60, 80, 100\}$
	Compute quantiles: $Q_i = P_{q_i}(\mathbf{X}), \quad i = \{0, 1, 2, 3, 4, 5\}$ where $P_{q}(\mathbf{X})$ is the q-th percentile of each column in $\mathbf{X}$.
	\item  Define a tensor $\mathbf{P}$ of shape $(m, 3, 4)$ to store partition parameters, and then compute partition parameters, as shown in Table \ref{tab:trapezoid_mem}.
	
	\begin{table}
		\centering
		\caption{Fuzzy membership trapezoid parameters}		
		\begin{tabular}{@{} l c c c c @{}}
			\toprule
			Partition & $P_{:,i,0}$ & $P_{:,i,1}$ & $P_{:,i,2}$ & $P_{:,i,3}$ \\
			\midrule
			Low$_{i=0}$ & $Q_0$ & $Q_0$ & $Q_1$ & $Q_2$ \\
			Medium$_{i=1}$ & $Q_1$ & $(Q_1 + Q_2)/2$ & $(Q_2 + Q_3)/2$ & $Q_3$ \\
			High$_{i=2}$ & $Q_2$ & $Q_3$ & $Q_4$ & $Q_4$ \\
			\bottomrule
		\end{tabular}
		\label{tab:trapezoid_mem}
	\end{table}
	
\end{enumerate}

\begin{figure}
	\centering
	\includegraphics[width=0.7\linewidth]{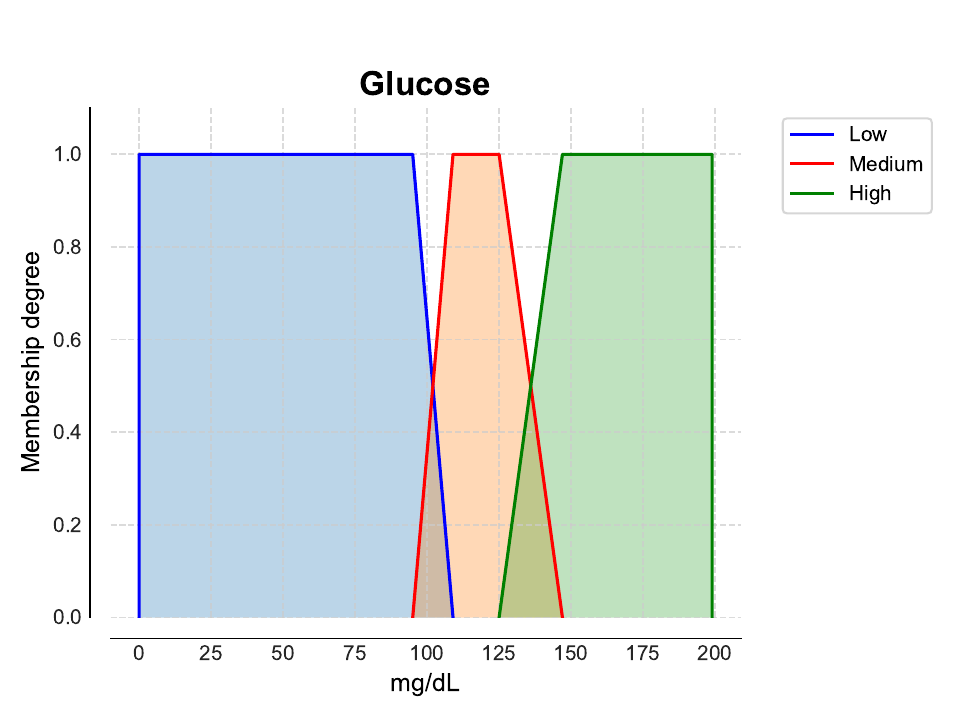}
	\caption{Visualization of fuzzy partitions using the method proposed for the glucose level measured during a 2-hour oral glucose tolerance test in the  \textit{Pima Indians Diabetes Dataset}.}
	\label{fig:temperature_fuzzy}
\end{figure}

The resulting tensor $\mathbf{P}$ contains the trapezoidal parameters for each variable and partition. As a visual example, Figure \ref{fig:temperature_fuzzy} shows a fuzzy partition computed this way for the classical Iris dataset.

\section{Example of a rule base obtained with the FRR}\label{apx:example}
As a way of illustration, Figure \ref{fig:diabetes_rules} shows an example of a rule base obtained with the FRR, for the \emph{Pima Indians Diabetes} dataset. This dataset contains diagnostic measurements from 768 adult female patients of Pima Native American heritage. It has 8 features:

\begin{itemize}
	\item Number of times pregnant.
	\item Glucose: Plasma glucose concentration (2-hour oral glucose tolerance test).
	\item Diastolic blood pressure (mm Hg).
	\item Triceps skin fold thickness (mm).
	\item Insulin: 2-hour serum insulin ($\mu$U/ml).
	\item BMI: Body mass index (weight in kg/(height in m)$^2$).
	\item Diabetes likelihood genetic score.
	\item Age in years.
\end{itemize}

Here we can see that the rules make clinical sense: a high glucose level is one of the most important indicators of diabetes \citep{american20212}. Negative indicators are also sound: a young person with low diabetes pedigree in his/her family is unlikely to develop diabetes (type 2), just as a person with low Body Mass Index. Number of pregnancies and skin thickening are other factors also used for diabetes diagnosis \citep{lima2017cutaneous}. 

%\begin{wrapfigure}{l}{0.55\textwidth}
\begin{figure}
	\centering
	\begin{tabular}{l}
		\toprule
		\multicolumn{1}{c}{\textbf{Rules for Non-Diabetic Patients}} \\
		\midrule
		\rowcolor{gray!25} IF Diabetes Pedigree IS Low \textbf{AND} Age IS Low \\
		IF Skin Thickness IS Low \textbf{AND} Insulin IS Medium \\
		\rowcolor{gray!25} IF Body Mass Index (BMI) IS Low \\
		IF Blood Pressure IS High \textbf{AND} Insulin IS Low \\
		\rowcolor{gray!25} IF Times Pregnant IS Medium \textbf{AND} \\
		\rowcolor{gray!25} Blood Pressure IS High \textbf{AND} Age IS Medium \\
		\midrule
		\multicolumn{1}{c}{\textbf{Rules for Diabetic Patients}} \\
		\midrule
		\rowcolor{gray!25} IF Glucose Level IS High \\
		\bottomrule
	\end{tabular}
	\caption{Classification Rules for the \textit{Pima Indians Diabetes Dataset} }%\citep{smith1988using}.}
\label{fig:diabetes_rules}
\end{figure}
%\end{wrapfigure}

\section{FRR component derivation} \label{apx:diff}

To optimize the cross-entropy loss we consider gradient-based optimization, so the parameters are going to be updated according to
\begin{equation}
	\mathbf{W}^{(l)}\mid_{t+1} = \mathbf{W}^{(l)}\mid_{t} -\eta_t \frac{\partial \mathcal{L}}{\partial \mathbf{u}^{(4)}} \cdot \frac{\partial 
		\mathbf{u}^{(4)}}{\partial \mathbf{W}^{(l)}},
\end{equation}

where $\eta_t$ is the learning rate. Then, to ensure an effective update of the model's parameters, let us analyse the derivative of each node with respect to its directly connected weights and nodes.

First of all, since all the weights are normalized using the softmax function (see Eq. (\ref{eq:softmax})), the derivative of each node according to the weights is multiplied by the derivative of the softmax, i.e, 
$$\frac{\partial u^{(l)}}{ \partial W_{i,j}^{(l)}} = \frac{\partial u^{(l)}}{ \partial \tilde{W}_{i,j}^{(l)}} \cdot \frac{\partial \tilde{W}_{i,j}^{(l)}}{\partial W_{i,j}^{(l)}},
$$
where 
\begin{equation}\label{eq:d1}
	\frac{\partial \tilde{W}_{i,j}^{(l)}}{\partial W_{i,m}^{(l)}}
	=
	\frac{1}{\alpha}\tilde{W}_{i,j}^{(l)}\cdot(\delta_{j,m}-\tilde{W}_{i,m}^{(l)}),
\end{equation}
and $\delta_{j,m}$ is the Kronecker delta. Then, we can compute the derivative of Eqs. (\ref{eq:u2}) and (\ref{eq:u3}) with respect to the normalized weights directly.

\begin{equation}\label{eq:d2}
	\frac{\partial u_j^{(2)}}{\partial \tilde{W}_{j,v}^{(2)}} =  \left(\frac{\partial f}{\partial \tilde{W}_{j,v}^{(2)}}\tilde{W}_{j,v}^{(2)}+f(\tilde{W}_{j,v}^{(2)}) \right) u_{j+v}^{(1)}.
\end{equation}
\begin{equation}\label{eq:d3}
	\frac{\partial u_j^{(2)}}{\partial u_{j+v}^{(1)}} = f(\tilde{W}_{j,v}^{(2)})\tilde{W}_{j,v}^{(2)}.
\end{equation}
\begin{eqnarray}\label{eq:d4}
	\frac{\partial u^{(3)}}{\partial \tilde{W}_{k,j}^{(3)}}
	&=& \left(\frac{\partial f}{\partial \tilde{W}_{k,j}^{(3)}}\tilde{W}_{k,j}^{(3)}+f(\tilde{W}_{k,j}^{(3)}) \right) u_{j}^{(2)} \cdot \nonumber \\
	&& \prod_{\substack{1 \leq m \leq A \\ m \not = k}} \sum_{n=1}^M f(\tilde{W}^{(3)}_{m,n}) \tilde{W}^{(3)}_{m,n}u_{n}^{(2)}.
\end{eqnarray}
\begin{equation}\label{eq:d5}
	\frac{\partial u^{(3)}}{\partial u_j^{(2)}} = \sum_{k=1}^A f(\tilde{W}_{k,j}^{(3)})\tilde{W}_{k,j}^{(3)} \prod_{\substack{1 \leq m \leq A \\ m \not = k}} \sum_{n=1}^M f(\tilde{W}^{(3)}_{m,n}) \tilde{W}^{(3)}_{m,n}u_{n}^{(2)}.
\end{equation}

%We omit the derivatives of Eqs. (\ref{eq:sufficient}) and (\ref{eq:additive}) since the last layer is a standard classification layer.

%To compute the derivatives of the fourth and last layer in which the classification step is performed we have to distinguish between the cases of sufficient and additive rules:
%\begin{itemize}
%	\item \textit{Sufficient rules}: 

Next, for simplifying the notation we consider $h_s(W_{s,c}^{(4)},r_s) = f(W_{s,c}^{(4)})W_{s,c}^{(4)}r_s(X_i)$, then $u^{(4)}_c = \max_{s \in \{1,\dots,R\}} h_s(W_{s,c}^{(4)},r_s)$ and we have
	\begin{equation}\label{eq:du4:r}
		\frac{\partial u^{(4)}_c}{\partial r_s} =    \left\{
		\begin{array}{ll}
			f(W_{s,c}^{(4)})W_{s,c}^{(4)} & \text{if } s = \argmax_k h_k,\\[0.2cm]
			0 & \text{otherwise.} 
		\end{array}
		\right.
	\end{equation}
	\begin{equation}\label{eq:du4:w}
		\frac{\partial u^{(4)}_c}{\partial W_{s,c}^{(4)}} = \left\{
		\begin{array}{ll}
			\left(\frac{\partial f}{\partial W_{s,c}^{(4)}}W_{s,c}^{(4)} + f(W_{s,c}^{(4)})\right) r_s(X_i) & \text{if } s = \argmax_k h_k,\\[0.2cm]
			0 & \text{otherwise.} 
		\end{array}
		\right.
	\end{equation}
	%Eq. (\ref{eq:du4:r}) is 0 whenever $W_{s,c}^{(4)}$ is not the biggest score, which blocks the gradient of the corresponding rule, and in Eq. (\ref{eq:du4:w})  we have to use a gradient estimator since $f$ is not differentiable. 
%	\item \textit{Additive rules}:
%	\begin{equation}
%		\frac{\partial u^{(4)}_c}{\partial r_s} = W_{s,c}^{(4)}, \quad \frac{\partial u^{(4)}_c}{\partial W_{s,c}^{(4)}} = r_s(X_i).
%	\end{equation}
%	In this case the derivatives correspond to the score and the truth degree, respectively.
%\end{itemize}

These derivatives visibly expose the issues in Section \ref{sec:training}, as it is now clear in which steps the derivative of the argmax function has to be approximated (see Eqs. (\ref{eq:d2}), (\ref{eq:d4}) and (\ref{eq:d6})) or where the gradient flow is blocked when the corresponding weight is not the biggest one (see Eqs. (\ref{eq:d3}) and (\ref{eq:d5})).

Next, we study how the different complementary components of the FRR affect the gradient-based optimization. First, we compute the derivatives involved in the update of the parameters in charge of reducing the number of conditions in each rule:% (see Eq. (\ref{eq:antecedent_silencer})):
\begin{equation}\label{eq:d6}
	\frac{\partial \tilde{A}_k}{\partial A_k} = f(\alpha_{k,1}) \prod_{\substack{1 \leq m \leq A \\ m \not = k}} f(\alpha_{m,1})A_m + f(\alpha_{m,2}).
\end{equation}
\begin{equation}\label{eq:d7}
	\frac{\partial \tilde{A}_k}{\partial \alpha_{k,1}} = \frac{\partial f}{\partial \alpha_{k,1}} A_k \prod_{\substack{1 \leq m \leq A \\ m \not = k}} f(\alpha_{m,1})A_m + f(\alpha_{m,2}).
\end{equation}
\begin{equation}\label{eq:d8}
	\frac{\partial \tilde{A}_k}{\partial \alpha_{k,2}} = \frac{\partial f}{\partial \alpha_{k,2}}\prod_{\substack{1 \leq m \leq A \\ m \not = k}} f(\alpha_{m,1})A_m + f(\alpha_{m,2}).
\end{equation}
These derivatives have a similar behaviour that the ones already discuss because of the presence of function $f$. Next, the derivative of the projection function is taken into account when the third layer is modified as in Eq. (\ref{eq:P}):
\begin{equation}
	\frac{dP}{dx} = \frac{1}{n\sqrt[n]{x^{n-1}}}.    
\end{equation}
Finally, the derivative of the third layer after including the residual connection is modified additively by the sum of derivatives of the conditions per rule weighted by $\gamma$:
\begin{equation}
	\frac{\partial \tilde{u}^{(3)}}{\partial u^{(3)}} = \frac{\partial u_j^{(3)}}{\partial u_j^{(2)}} + \gamma \sum_{k=1}^A \frac{\partial A_k}{\partial u_j^{(2)}} = \frac{\partial u_j^{(3)}}{\partial u_j^{(2)}} + \gamma \sum_{k=1}^A f(W_{k,j}^{(3)})W_{k,j}^{(3)}.
\end{equation}
\begin{equation}
	\frac{\partial \tilde{u}^{(3)}}{\partial W^{(3)}_{k,j}} = \frac{\partial u_j^{(3)}}{\partial W^{(3)}_{k,j}} + \gamma \sum_{k=1}^A \frac{\partial A_k}{\partial W^{(3)}_{k,j}} = \frac{\partial u_j^{(3)}}{\partial W_{k,j}^{(3)}} + \gamma \sum_{k=1}^A \left(\frac{\partial f}{\partial W_{k,j}^{(3)}}W_{k,j}^{(3)}+f(W_{k,j}^{(3)})\right)u_{j}^{(2)}.
\end{equation}

\section{Using the FRR with other conjunctions}\label{apx:tnorms}
In fuzzy logic, T-norms are considered to be the extension of the boolean conjunction and are defined as binary functions $T:[0,1]^2 \to [0,1]$ which are commutative, associative, increasing in both variables, and 1 is its neutral element (i.e., $T(x,1)=x$ for all $x \in [0,1]$). Since the properties imposed in this definition are not very restrictive, there exist a lot of functions that can be used to model fuzzy conjunctions. However, in practice, the minimum $T_M(x,y)=\min\{x,y\}$ and the product $T_P(x,y)=x \cdot y$ are the preferred choice because they are easy to implement and have a simple $n$-ary form:
$$T_M(x_1,\dots,x_n) = \min \{x_1,\dots,x_n\}, \quad T_P(x_1,\dots,x_n) = \prod_{i=1}^nx_i.$$
In general, since all T-norms are associative, the order of the inputs does not change the output, and any t-norm defines an $n$-ary function which is equivalent to the iterative evaluation of each instance.
$$
T(x_1,T(x_2,\dots T(x_{n-1},x_n)))=T(x_1,\dots,x_n) = \underset{i=1}{\overset{n}{\mathbf{\text{\large T}}}}x_i.
$$
However, no T-norm has an easy closed expression of its $n$-ary form, and that is one of the reasons why the minimum and the product are mostly used. Nonetheless, continuous Archimedean T-norms are a special type that can be constructed via a unary continuous, strictly decreasing function $t:[0,1] \to [0,+\infty)$ with $t(1)=0$ called generator and they have a closed $n$-ary expression that allows a more efficient implementation \citep{Giannini2023}:
$$
T(x_1,\dots,x_n) = t^{-1}\left(\min\left\lbrace t(0^+),\sum_{i=1}^nt(x_i)\right\rbrace\right).
$$
Any of these T-norms can be used in the FRR instead of the product to combine weights with inputs and to perform the logic inference during the second and third layers in the following manner
$$
u_j^{(2)}= \sum_{v=1}^V T(f(\tilde{W}^{(2)}_{j,v}) \tilde{W}^{(2)}_{j,v},\mu_{j,v}(X_{i,j})),
$$
$$u^{(3)} = 
\underset{k=1}{\overset{A}{\mathbf{\text{\large T}}}} \sum_{j=1}^M T(f(\tilde{W}^{(3)}_{k,j}) \tilde{W}^{(3)}_{k,j},u_{j}^{(2)}) = \underset{k=1}{\overset{A}{\mathbf{\text{\large T}}}} \sum_{j=1}^M T(f(\tilde{W}^{(3)}_{k,j}) \tilde{W}^{(3)}_{k,j},\sum_{v=1}^V T(f(\tilde{W}^{(2)}_{j,v}) \tilde{W}^{(2)}_{j,v},\mu_{j,v}(X_{i,j}))).$$
If we resolve the argmax function and we take into account that $T(x,0)=0$ for all $x \in [0,1]$ the truth degree of a rule $r$ is given by
$$
r(X_i) = \underset{k=1}{\overset{A}{\mathbf{\text{\large T}}}} T(\tilde{W}^{(3)}_{k,j_k},T(\tilde{W}^{(2)}_{j_k,v_{j_k}}\mu_{j_k,v_{j_k}}(X_{i,j_k}))) = \underset{k=1}{\overset{A}{\mathbf{\text{\large T}}}} T(T(\tilde{W}^{(3)}_{k,j_k},\tilde{W}^{(2)}_{j_k,v_{j_k}}),\mu_{j_k,v_{j_k}}(X_{i,j_k})).
$$

\begin{figure}[ht!]
	\centering
	% First Subfigure
	\begin{subfigure}[b]{0.3\textwidth}
		\centering
		\includegraphics[width=\textwidth]{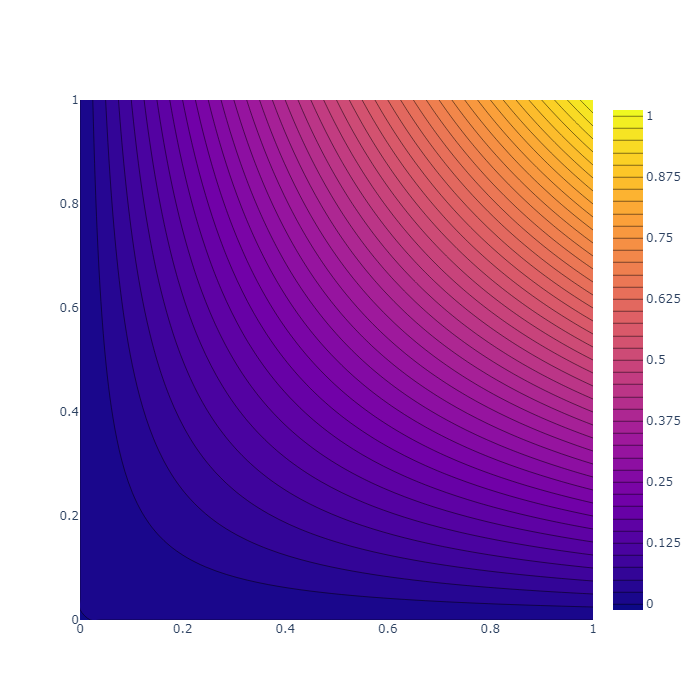} % Replace with your image file
		\caption{$T_{1}$}
	\end{subfigure}
	\hfill
	% Second Subfigure
	\begin{subfigure}[b]{0.3\textwidth}
		\centering
		\includegraphics[width=\textwidth]{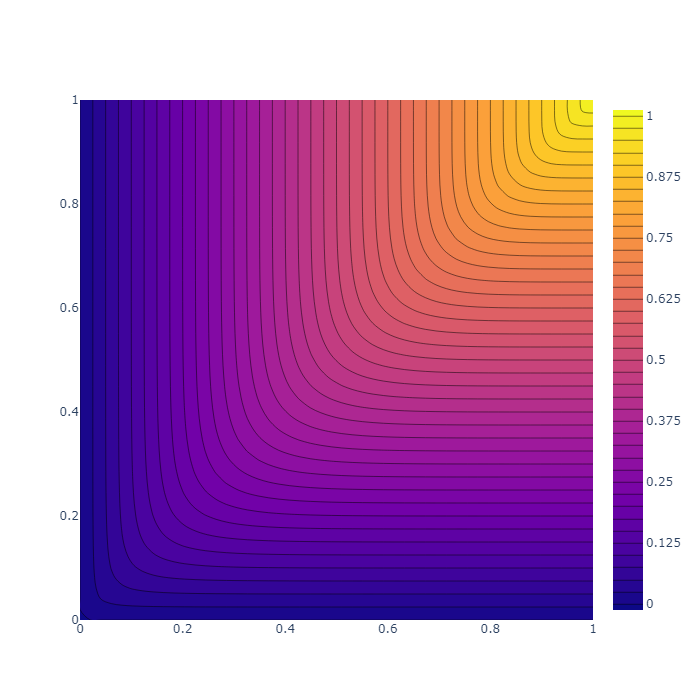} % Replace with your image file
		\caption{$T_{5}$}
	\end{subfigure}
	\hfill
	% Third Subfigure
	\begin{subfigure}[b]{0.3\textwidth}
		\centering
		\includegraphics[width=\textwidth]{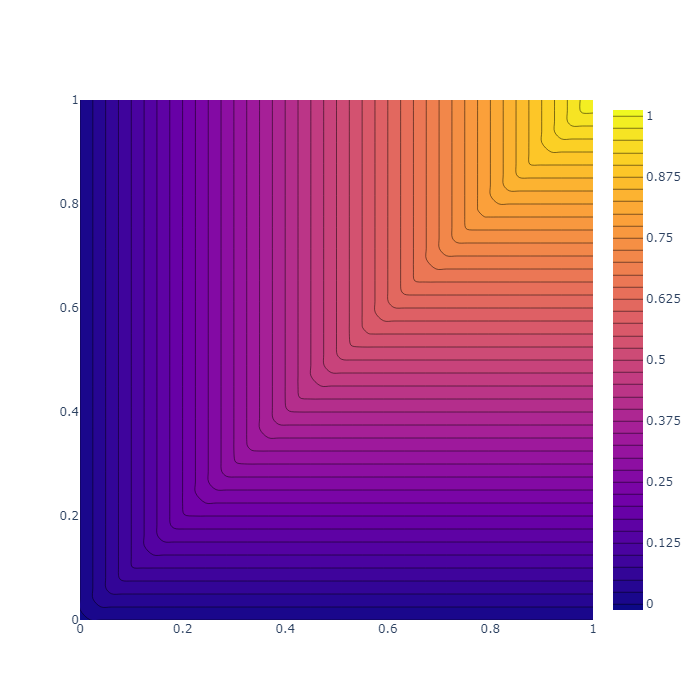} % Replace with your image file
		\caption{$T_{100}$}
	\end{subfigure}
	
	\caption{Contour plots of the Aczél-Alsina t-norm for different values of the parameter $\lambda$.}
	\label{fig:aa_tnorms}
\end{figure}

By definition, any T-norm is below the minimum $T(x,y) \leq \min \{x,y\}$, so $T_M$ is the greatest T-norm. Then, using the minimum t-norm is the best choice to not accelerate the approach to 0 when doing the logical inference. However, the minimum only takes into account the least value of all the inputs, which may not be desirable in some cases because it may happen that the truth degrees of the fuzzy sets are neglected. Even so, there exist other families of continuous differentiable T-norms that are above the product that may help in the vanishing gradient problem, like, for instance, the family of Aczél-Alsina T-norms:
$$t_{\lambda}(x) = (-\log x)^{\lambda}, \quad T_{\lambda}(x_1,\dots,x_n)= e^{-\sqrt[\lambda]{\sum_{i=1}^n(-\log x_i)^{\lambda}}},$$
with $\lambda \in [1,+\infty)$. As $\lambda \to +\infty$, the T-norm converges to the minimum and if $\lambda=1$ is equal to the product (see Figure \ref{fig:aa_tnorms}). Also, the derivatives of different logic operators can be interpreted in terms of their contribution to the gradient \citep{Vankrieken2022}. However, the use of an arbitrary T-norm may affect the explainability of the system.

\section{Reducing the size of the FRR} \label{sec:reduce_complexity}
Even though we can set up a maximum number of rules and conditions per rule, there are sometimes cases where the FRR can find good solutions that are considerably smaller than these specifications. In our experimentation, we found some modifications in the loss function that helped reduce the size of the rules and their number.

The strategy to shorten the rules is to add a Laplacian term in the loss function that penalizes the weights in the cancellation process that affect the condition truth value:
\begin{equation}
	L_{cancellation} =  \sum_{r=1}^{R} \sum_{k=1}^{A} \alpha_{k,1}^{(r)},
\end{equation}

which is then added to the cross-entropy loss with a multiplier, which in our experimentation was set to $0.01$.

\section{Datasets used and Code availability} \label{sec:datasets}

\begin{table}[ht]
	\centering
	\caption{Datasets with their samples, features, classes, and categorical variables.}
	\begin{tabular}{l|rrrr}
		\hline
		\textbf{Dataset} & \textbf{Samples} & \textbf{Features} & \textbf{Classes} & \textbf{Categorical} \\ 
		\midrule
		appendicitis     & 106             & 7                & 2               & 0                   \\ 
		australian       & 690             & 14               & 2               & 4                   \\ 
		balance          & 625             & 4                & 3               & 0                   \\ 
		banana           & 5300            & 2                & 2               & 0                   \\ 
		bands            & 512             & 39               & 2               & 5                   \\ 
		bupa             & 345             & 6                & 2               & 0                   \\ 
		cleveland        & 297             & 13               & 5               & 3                   \\ 
		coil2000         & 9822            & 85               & 2               & 10                  \\ 
		dermatology      & 358             & 34               & 6               & 2                   \\ 
		glass            & 214             & 9                & 6               & 0                   \\ 
		haberman         & 306             & 3                & 2               & 0                   \\ 
		heart            & 270             & 13               & 2               & 3                   \\ 
		hepatitis        & 80              & 19               & 2               & 4                   \\ 
		housevotes       & 232             & 16               & 2               & 16                  \\ 
		ionosphere       & 351             & 33               & 2               & 0                   \\ 
		iris             & 150             & 4                & 3               & 0                   \\ 
		magic            & 19020           & 10               & 2               & 0                   \\ 
		mammographic     & 830             & 5                & 2               & 2                   \\ 
		monk-2           & 432             & 6                & 2               & 6                   \\ 
		newthyroid       & 215             & 5                & 3               & 0                   \\ 
		page-blocks      & 5472            & 10               & 5               & 0                   \\ 
		phoneme          & 5404            & 5                & 2               & 0                   \\ 
		pima             & 768             & 8                & 2               & 0                   \\ 
		ring             & 7400            & 20               & 2               & 0                   \\ 
		saheart          & 462             & 9                & 2               & 2                   \\ 
		satimage         & 6435            & 36               & 6               & 0                   \\ 
		sonar            & 208             & 60               & 2               & 0                   \\ 
		spambase         & 4597            & 57               & 2               & 0                   \\ 
		spectfheart      & 267             & 44               & 2               & 0                   \\ 
		thyroid          & 7200            & 21               & 3               & 3                   \\ 
		titanic          & 2201            & 3                & 2               & 2                   \\ 
		twonorm          & 7400            & 20               & 2               & 0                   \\ 
		vehicle          & 846             & 18               & 4               & 0                   \\ 
		wdbc             & 569             & 30               & 2               & 0                   \\ 
		wine             & 178             & 13               & 3               & 0                   \\ 
		winequality-red  & 1599            & 11               & 6               & 0                   \\ 
		winequality-white & 4898           & 11               & 7               & 0                   \\ 
		wisconsin        & 683             & 9                & 2               & 0                   \\ 
		zoo              & 101             & 16               & 7               & 15                  \\ 
		\bottomrule
	\end{tabular}
	\label{tab:uci_datasets}
\end{table}

The list of datasets used, alongside the number of features, samples, and classes, is in Table \ref{tab:uci_datasets}. They were collected from the UCI datasets \citep{kelly2025uci} and the Keel website \citep{triguero2017keel}. The code will be publicly available on Github. 

The following systems were implemented using their public corresponding code: FGA (implemented using the exFuzzy library) \citep{fumanalex2024}, RRL \citep{wang2021scalable}, DRNet \citep{qiao2021learning} and GOT \citep{mctavish2022fast}. For RIPPER, we used a public repository and then we used a OVR scheme. We also considered using \citep{yang2024hyperlogic}, but the results obtained using any of the configuration provided by the authors in their code were very underperforming classifiers in most datasets. C4.5 classifier and SIRUS were implemented in Python for this work. Baseline models (linear regression and gradient boosting) were adopted from \citep{scikit-learn}. All code repositories are publicly available:
\begin{itemize}
	\item exFuzzy: \url{https://github.com/Fuminides/ex-fuzzy}
	\item RRL: \url{https://github.com/12wang3/rrl}
	\item DRNet: \url{https://github.com/Joeyonng/decision-rules-network}
	\item HyperNet: \url{https://github.com/YangYang-624/HyperLogic}
	\item GOT: \url{https://github.com/ubc-systopia/gosdt-guesses/tree/main}
	\item RIPPER: \url{https://github.com/imoscovitz/wittgenstein}.
\end{itemize}

%\section{Computational resources employed} \label{sec:exp_setting}

%For the experiments using gradient-based optimization, we used the CERES cluster from the University of Essex. CERES has 1096 processing cores (2192 with hyperthreading) provided by servers with a mix of Intel E5-2698, Intel Gold 5115, 6152 and 6238L processors, and between 500Gb and 6Tb RAM each. There are also 24 NVidia GTX and RTX Series GPU cards (16 x GTX1080Ti and 8 x RTX2080). Figure \ref{fig:time_hist} shows the execution time for all datasets for $400$ epochs with the FRR. The average time was 218 seconds.

%For the rest of the experiments, we used the Oracle Cloud Infrastructure. We set up two machines with 128GB of RAM and 32 cores. 
%\begin{figure}
%	\centering
%	\includegraphics[width=0.5\linewidth]{Time_histogram.pdf}
%	\caption{Histogram with execution times for all datasets using the compact FRR}
%	\label{fig:time_hist}
%\end{figure}

\section{Hyperparameter choosing and detailed performance for each method} \label{sec:detailed_res}

We tried different configurations of hyperparameters for all the classifiers tested. The ones reported in the main text are the ones that achieved the best accuracy results for each of them.

\begin{itemize}
	\item Gradient boosting: we tried $100$ and $200$ number of trees.
	\item CART: we tried trees with three different cost complexity pruning parameters. The higher this parameter, the smaller the final tree shall be. We tried: $0.0$, $0.001$ and $0.003$.
	\item RRL: for the RRL, we used the configurations that the authors recommend in \citep{wang2021scalable}. However, we obtained very similar results in all cases.
	\item C4.5: we tried a maximum depth of $5$ and $10$.
	\item DINA: we tried the different configurations that the authors reported in their public codes. They used different ones for each of their examples, so we runned them all and took the best. We lowered the batch size from the recommended value ($400$) to $32$, as it increased performance in most cases. We trained the models for $1000$ epochs.
	\item DRNet: same as DINA.
	\item GOT: we tried a maximum depth of $3$ and $5$.
	\item SIRUS: results are reported with $100$ stimators and a maximum depth of $3$.
	\item FGA: results are reported in the main text of the work with $15$ maximum rules and $3$ fuzzy sets per rule. We also tried configurations with up to $100$ maximum rules and up to $5$ fuzzy sets per rule, but obtained worse results than the smaller configurations.
	\item RIPPER: parameters were set based on the defaults of the Wittgenstein repository.
\end{itemize}

\begin{table*}[ht]
	\centering
	\caption{Results per each dataset obtained with the best configuration per each classifier.}
	\begin{adjustbox}{width=\linewidth}
		\begin{tabular}{lrrrrrrrrrrrr}
			\toprule
			Dataset            & FRR    &  DINA&SIRUS &DRNet& GOT    & LR     & GB     & FGA     & RLL    & CART   & C4.5   \\
			\midrule                    
                appendicitis       &  86.36 & 80.00&81.81 &77.27& 77.27  & 86.36  & 84.55  & 90.91  & 84.03  & 79.09  & 94.02  \\
			australian         &  83.91 & 59.42&78.98 &56.52& 84.78  & 86.23  & 86.81  & 85.51  & 86.67  & 85.07  & 95.46  \\
			balance            &  79.20 & 51.61&88.00 &46.40& 69.60  & 88.16  & 84.16  & 61.60  & 84.48  & 78.56  & 73.00  \\
			banana             &  78.36 & 53.39&88.58 &57.64& 67.36  & 55.70  & 89.51  & 74.06  & 77.36  & 89.04  & 55.42  \\
			bands              &  67.12 & 58.33&75.34 &36.98& 71.23  & 67.12  & 72.88  & 64.38  & 61.64  & 63.56  & 87.55  \\
			bupa               &  70.14 & 55.88&66.66 &43.47& 68.12  & 68.12  & 75.36  & 55.07  & 65.80  & 65.51  & 66.81  \\
			cleveland          &  58.00 & 34.48&53.33 &53.33& 58.33  & 58.00  & 56.00  & 46.67  & 54.22  & 50.00  & 96.82  \\
			coil2000           &  92.61 & 94.90&92.97 &94.04& 94.05  & 93.98  & 92.81  & 80.81  & 93.33  & 94.05  & 98.90  \\
			dermatology        &  80.05 & 25.71&97.22 &30.55& 58.33  & 95.56  & 97.22  & 50.00  & 95.53  & 94.72  & 56.05  \\
			glass              &  63.23 & 23.80&62.79 &32.55& 55.81  & 62.33  & 73.49  & 39.53  & 65.87  & 66.98  & 100.00 \\
			haberman           &  62.90 & 36.66&74.19 &40.32& 72.58  & 76.45  & 70.65  & 50.00  & 72.23  & 67.42  & 56.54  \\
			heart              &  80.03 & 59.25&81.48 &55.55& 81.48  & 83.70  & 82.96  & 70.37  & 84.81  & 73.70  & 66.17  \\
			hepatitis          &  90.00 & 50.00&93.75 &12.50& 93.75  & 85.00  & 88.75  & 81.25  & 80.00  & 82.50  & 75.89  \\
			housevotes         &  80.01 & 65.21&95.74 &95.74& 95.74  & 96.17  & 97.02  & 97.87  & 95.70  & 97.87  & 100.00 \\
			ionosphere         &  85.07 & 57.14&92.95 &63.38& 90.14  & 87.61  & 92.39  & 78.87  & 92.01  & 85.63  & 100.00 \\
			iris               &  96.66 & 53.33&96.66 &50.06& 66.67  & 95.33  & 96.00  & 76.67  & 95.33  & 96.67  & 62.50  \\
			magic              &  77.81 & 71.03&86.30 &64.82& 81.52  & 79.02  & 88.23  & 76.37  & 71.39  & 83.17  & 65.08  \\
			mammographic       &  82.77 & 79.51&78.31 &51.20& 80.12  & 81.33  & 80.12  & 77.71  & 79.88  & 82.65  & 86.04  \\
			monk-2             &  95.17 & 48.83&00.00 &49.42& 100.00 & 78.62  & 98.85  & 63.22  & 100.00 & 100.00 & 97.10  \\
			newthyroid         &  93.48 & 14.28&95.34 &13.95& 79.07  & 94.88  & 96.28  & 88.37  & 94.88  & 93.49  & 79.56  \\
			page-blocks        &  91.01 & 84.09&96.98 &89.68& 94.25  & 95.95  & 97.37  & 92.60  & 94.63  & 96.44  & 94.74  \\
			phoneme            &  78.33 & 71.29&84.82 &70.67& 78.72  & 75.00  & 90.36  & 67.90  & 83.25  & 82.35  & 71.31  \\
			pima               &  72.59 & 61.84&76.62 &64.93& 79.22  & 77.66  & 77.14  & 74.68  & 74.08  & 71.95  & 69.24  \\
			ring               &  84.41 & 55.00&93.10 &49.45& 73.31  & 75.64  & 95.00  & 68.58  & 91.96  & 85.88  & 51.98  \\
			saheart            &  70.96 & 47.82&64.51 &65.59& 77.42  & 76.13  & 68.17  & 66.67  & 67.30  & 60.86  & 84.74  \\
			satimage           &  80.32 & 51.16&87.87 &23.77& 70.94  & 85.58  & 91.02  & 61.07  & 89.29  & 83.98  & 76.12  \\
			sonar              &  80.00 & 55.00&83.33 &47.61& 73.81  & 77.14  & 83.81  & 61.90  & 58.65  & 71.43  & 100.00 \\
			spambase           &  78.78 & 66.23&94.13 &54.56& 90.43  & 91.91  & 95.20  & 88.48  & 91.45  & 91.02  & 99.35  \\
			spectfheart        &  77.40 & 65.38&88.88 &20.37& 81.48  & 81.11  & 81.85  & 64.81  & 79.01  & 74.44  & 100.00 \\
			thyroid            &  92.56 & 71.11&99.02 &02.91& 94.93  & 95.65  & 99.64  & 88.54  & 94.51  & 99.39  & 94.01  \\
			titanic            &  75.96 & 67.27&80.95 &74.60& 78.46  & 77.64  & 78.91  & 78.91  & 77.96  & 78.37  & 78.26  \\
			twonorm            &  94.47 & 51.75&95.33 &50.00& 75.27  & 97.50  & 96.96  & 79.73  & 95.22  & 79.76  & 51.18  \\
			vehicle            &  62.27 & 26.19&78.23 &23.52& 64.71  & 78.82  & 74.82  & 52.94  & 71.52  & 68.12  & 79.62  \\
			wdbc               &  94.91 & 42.85&95.61 &48.24& 93.86  & 97.54  & 96.49  & 89.47  & 95.26  & 94.04  & 84.61  \\
			winequality-red    &  54.62 & 29.41&94.44 &33.33& 43.75  & 59.56  & 67.62  & 23.12  & 59.66  & 59.31  & 52.39  \\
			winequality-white  &  49.91 & 28.30&63.75 &00.62& 44.90  & 53.73  & 66.47  & 20.61  & 54.10  & 52.47  & 50.12  \\
			wine               &  97.22 & 32.10&53.97 &00.40& 83.33  & 98.89  & 97.78  & 91.67  & 96.60  & 92.22  & 70.79  \\
			wisconsin          &  97.51 & 92.64&97.81 &64.96& 97.81  & 96.50  & 96.93  & 95.62  & 95.32  & 94.74  & 98.62  \\
			zoo                &  66.85 & 40.00&95.23 &57.14& 57.14  & 91.43  & 96.19  & 71.43  & 93.00  & 95.24  & 100.00 \\
			\bottomrule                  
		\end{tabular}
	\label{tab:classical_results}
\end{adjustbox}
\end{table*}
For the number of epochs, we went with $300$, which was probably more than needed in most cases but did not require a significant time investment in most cases. 
%Two examples of the evolution of loss and accuracy in training and validation, respectively, are shown in Figure \ref{fig:training_loss}. We can see there how the loss in training remains stable until the last epochs, where the $k$ and $\gamma$ values approach 0.

Detailed results for every classifier tested are available in Table \ref{tab:classical_results}.
%\begin{figure}[ht]
%	\centering
%	\begin{subfigure}[b]{0.48\linewidth}
%		\centering
%		\includegraphics[width=\linewidth]{Example loss evolution australian.pdf}
%		\caption{Australian dataset}
%	\end{subfigure}
%	\hfill
%	\begin{subfigure}[b]{0.48\linewidth}
%		\centering
%		\includegraphics[width=\linewidth]{Example loss evolution magic.pdf}
%		\caption{Magic dataset}
%	\end{subfigure}
%	\caption{Performance evolution for the FRR during two different training sessions for two datasets.}
%	\label{fig:training_loss}
%\end{figure}

\newpage

\end{document}